%% file: Main.tex
\documentclass[10pt,twocolumn,letterpaper]{article}
\usepackage{3dv}
\usepackage{times}
\usepackage{epsfig}
\usepackage{graphicx}
\usepackage{amsmath}
\usepackage{amssymb}
\usepackage{booktabs}
\usepackage{makecell}
\usepackage{multirow}
\usepackage{csquotes}
\usepackage{caption}
\usepackage{subcaption}
\usepackage[pagebackref=true,breaklinks=true,colorlinks,bookmarks=false]{hyperref}
\threedvfinalcopy
\def\threedvPaperID{59} 

\ifthreedvfinal\fi
\begin{document}

\newcommand{\printfnsymbol}[1]{%
  \textsuperscript{\textasteriskcentered}%
}

\title{Investigating Attention Mechanism in 3D Point Cloud Object Detection}

\author{Shi Qiu\thanks{denotes equal contributions.} $^{,1,2}$, Yunfan Wu\printfnsymbol{1}$^{,1}$, Saeed Anwar$^{1,2,3}$ and Chongyi Li$^{4}$\\
 $^1$Australian National University, $^2$Data61-CSIRO, Australia\\ $^3$University of Technology Sydney, $^4$Nanyang Technological University\\
{\tt\small \{shi.qiu, yunfan.wu, saeed.anwar\}@anu.edu.au, chongyi.li@ntu.edu.sg}}

\maketitle
\thispagestyle{empty}
\begin{abstract}
Object detection in three-dimensional (3D) space attracts much interest from academia and industry since it is an essential task in AI-driven applications such as robotics, autonomous driving, and augmented reality. As the basic format of 3D data, the point cloud can provide detailed geometric information about the objects in the original 3D space. However, due to 3D data's sparsity and unorderedness, specially designed networks and modules are needed to process this type of data. Attention mechanism has achieved impressive performance in diverse computer vision tasks; however, it is unclear how attention modules would affect the performance of 3D point cloud object detection and what sort of attention modules could fit with the inherent properties of 3D data. This work investigates the role of the attention mechanism in 3D point cloud object detection and provides insights into the potential of different attention modules. To achieve that, we comprehensively investigate classical 2D attentions, novel 3D attentions, including the latest point cloud transformers on SUN RGB-D and ScanNetV2 datasets. Based on the detailed experiments and analysis, we conclude the effects of different attention modules. This paper is expected to serve as a reference source for benefiting attention-embedded 3D point cloud object detection. The code and trained models are available at: \url{https://github.com/ShiQiu0419/attentions_in_3D_detection}.
\end{abstract}

\input{sections/intro}

\input{sections/related}

\input{sections/approach}

\input{sections/exp}

\input{sections/conclusion}

{\small
\bibliographystyle{ieee_fullname}
\bibliography{egbib}
}

\end{document}

%% file: sections/intro.tex
\section{Introduction}
3D data such as point cloud provides detailed geometric and structure information in scene understandings compared to images. As a result, the applications of 3D data have attracted more and more attention in recent years. Among the applications, 3D point cloud object detection is the most essential function, which is highly desired in robotics, autonomous driving, and augmented reality. Due to the irregular format of 3D point cloud data, the pipelines of 3D object detection~\cite{qi2018frustum, qi2019deep, qi2020imvotenet} differ from conventional 2D object detection methods~\cite{girshick2015fast, ren2015faster, he2017mask}, raising new challenges.

\begin{figure}
     \centering
     \begin{subfigure}[b]{\columnwidth}
         \centering
         \includegraphics[width=0.85\textwidth]{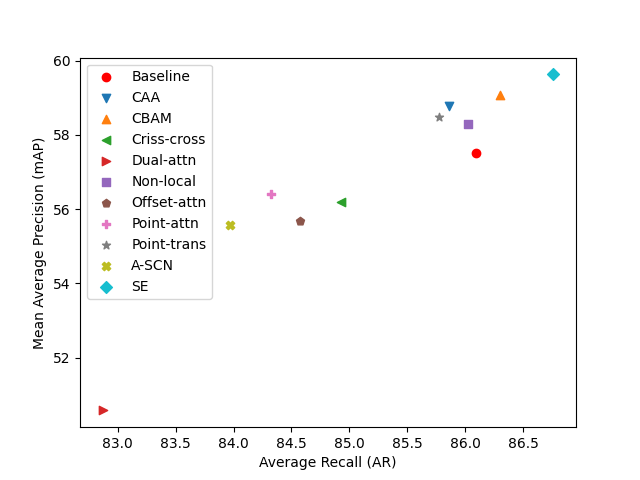}
         \caption{Overall results (\%) on \emph{SUN RGB-D}~\cite{song2015sun} dataset.}
     \end{subfigure}
     \hfill
     \begin{subfigure}[b]{\columnwidth}
         \centering
         \includegraphics[width=0.85\textwidth]{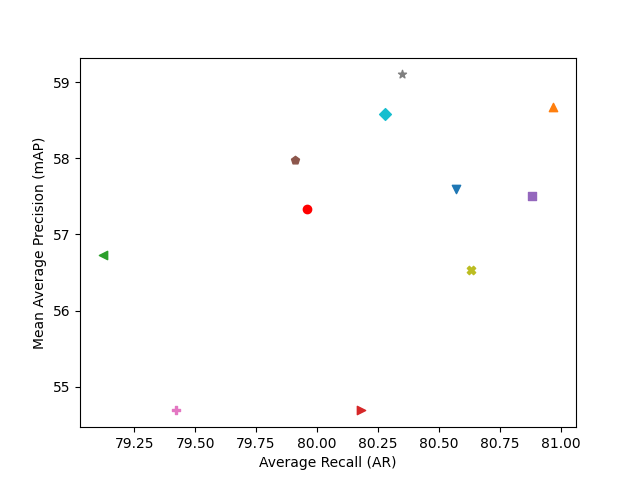}
         \caption{Overall results (\%) on \emph{ScanNetV2}~\cite{dai2017scannet} dataset.}
     \end{subfigure}
     \vspace{-5mm}
        \caption{The performances (IoU threshold=0.25) of using different attention modules in VoteNet~\cite{qi2019deep} backbone for 3D point cloud object detection.}
        \label{fig:sun}
\vspace{-5mm}
\label{fig:result}
\end{figure}

Many frameworks~\cite{qi2017pointnet, su2015multi, choy20194d, thomas2019kpconv} have been proposed to process the irregular point cloud data for object detection. For example, VoxNet~\cite{maturana2015voxnet} voxelized the point cloud data and used 3D CNNs to detect the objects, but it usually suffers from the expensive computation cost. To share the solid performance of 2D detectors, some methods~\cite{chen2017multi,ku2018joint} project the point cloud data to front view or bird's eye view (BEV) and lift the existing 2D detectors for 3D region proposals. However, as the occlusion problem is still challenging in 2D detection, this issue could also be problematic in 3D space.


PointNet \cite{qi2017pointnet} and PointNet++ \cite{qi2017pointnet++} are proposed to process point cloud data directly without voxelization or projection into BEVs. This track of methods achieves excellent performance in classification~\cite{qiu2021dense} and semantic segmentation~\cite{qiu2021semantic} tasks. Although these methods can effectively process the unordereded point cloud data, they cannot be directly used for 3D object detection due to point cloud's inherent sparsity. In 2D images, the center of an object is likely to be inside the object's pixels, while the center of an object in 3D point cloud is usually in empty space since only the surfaces of an object can be captured by the 3D scanners~\cite{blais2004review, jaboyedoff2012use}. Thus, it is hard to aggregate the context around an object center, which leads to difficulties in detecting 3D objects in the point cloud. 

To address this problem, VoteNet~\cite{qi2019deep} generates some seed points that are close to object centers via a PointNet++ backbone and then produces the region proposals after voting and clustering the seed points. Moreover, VoteNet is an end-to-end trainable network that significantly simplifies the 3D object detection pipeline by omitting 2D to 3D conversions and appropriate pre/post-processing steps. Although VoteNet is efficient in 3D object detection, its effectiveness highly relies on the point features learned from the backbone network. In detail, the learned point features would not only affect the selection of seed points, but also involve in the following voting process since the offsets from the object centers are totally estimated from them. Therefore, extracting high-quality point features is a critical factor for the success of such a 3D object detection pipeline. 

Feature learning is always considered as a fundamental problem in all computer vision tasks, especially in using Convolutional Neural Networks (CNNs). Following the success of the attention mechanism~\cite{vaswani2017attention} in language-related topics, attention as a basic function for extracting and refining features can significantly improve the performance of many computer vision tasks. Basically, regular attention modules used in image feature learning can be categorized into three main types based on the operating domains: spatial attention~\cite{wang2018non, huang2019ccnet}, channel attention~\cite{hu2018squeeze}, and mixed attention~\cite{woo2018cbam, fu2019dual}. More recently, an increasing number of attention modules are proposed for 3D point cloud analysis, including self-attention~\cite{xie2018attentional, feng2020point, qiu2021geometric} and transformer~\cite{guo2021pct, zhao2020point} methods. Although the attention mechanism shows the effectiveness in point cloud classification~\cite{qiu2021geometric} or semantic segmentation~\cite{yan2020pointasnl}, fewer of them are utilized in the point cloud object detection since it is unclear if the attention mechanism also fits with this task. As we explained above, the learned point features in VoteNet~\cite{qi2019deep} are critical for 3D object detection. Thus, taking the recent VoteNet~\cite{qi2019deep} as a basic pipeline, we investigate the advantages and disadvantages of different attention modules in 3D point cloud object detection.

To thoroughly investigate the attention mechanism's effects on 3D point cloud object detection, we analyze \emph{five} classical 2D attention modules~\cite{hu2018squeeze, wang2018non, huang2019ccnet, woo2018cbam, fu2019dual} and \emph{five} novel 3D attention modules~\cite{xie2018attentional, feng2020point, qiu2021geometric, guo2021pct, zhao2020point} (more details are provided in Section~\ref{sec:app}). Furthermore, we conduct the experiments on two widely-used 3D object detection benchmarks, \emph{SUN RGB-D}~\cite{song2015sun} and \emph{ScanNetV2}~\cite{dai2017scannet} datasets, under different metrics such as the ones shown in Figure~\ref{fig:result}. In general, our main contributions can be concluded from the following aspects:
\begin{itemize}
    \item We push the VoteNet pipeline towards better performance for 3D point cloud object detection by integrating attention mechanisms into it. 
    \item We are the first to comprehensively evaluate the performances of \emph{ten} recent attention modules for 3D point cloud object detection on \emph{SUN RGB-D} and \emph{ScanNetV2} datasets.
    \item We concretely summarize the effects and characters of different types of attention modules and provide novel insights and inspiration to facilitate the understanding of the attention mechanism for 3D point cloud object detection.
\end{itemize}

%% file: sections/related.tex
\section{Related Work}

\noindent \textbf{Point Cloud Networks.}
Due to the rapid development of 3D sensors, point cloud data can be easily collected using LiDAR scanners and RGB-D cameras. To better understand the contained information in point cloud data, different CNN-based point cloud networks have been invented for machine perception. To be specific, early methods~\cite{su2015multi, yu2018multi, maturana2015voxnet} attempt to convert raw point clouds into a particular intermediate representation (\eg, images or voxels) according to the projective relations, then apply regular 2D/3D CNN operations to learn the high-dimensional features for the following analysis. However, the intermediate representations of point cloud data usually encounter the problem of geometric information loss, resulting in inaccurate predictions.

To avoid this problem, current methods tend to exploit the usage of Multi-Layer Perceptron (MLP), which was originally proposed in PointNet~\cite{qi2017pointnet}. In practice, MLP is implemented as the $1\times 1$ convolutions followed by a Batch Normalization (BN) layer and an activation function. Particularly, MLP can avoid the sparsity and unorderedness of raw point cloud data because each point shares the learnable weights. Moreover, PointNet++~\cite{qi2017pointnet++} extends the basic MLP operation to aggregate the local features from pre-defined point neighborhoods via a symmetric function like max-pooling. Since VoteNet~\cite{qi2019deep} takes PointNet++ as the backbone for feature learning, we mainly use MLP as the essential convolutional operation in the attention mechanism for 3D point cloud object detection.  

\vspace{3mm}
\noindent \textbf{3D Object Detection.}
The aim of object detection in 3D space is to predict the class label and 3D bounding box for each object. In general, the standard approaches for 3D object detection can be categorized into two streams~\cite{guo2020deep}: \emph{the region proposal-based} and \emph{single-shot methods}. 

Region proposal-based methods are usually in two-stage where the first stage generates region proposals and the second stage decides the class label of each proposal. More concretely, the region proposal-based methods are in three tracks: the multi-view based methods~\cite{chen2017multi, ku2018joint, liang2018deep}, segmentation-based methods~\cite{shi2019pointrcnn, vora2020pointpainting, yang2019std}, and frustum-based methods~\cite{qi2018frustum, xu2018pointfusion, wang2019frustum}. In addition, the single-shot methods can be more efficient than the region proposal-based methods since they directly estimate the class probabilities and regress bounding boxes. According to the ways of processing the raw 3D data, the single-shot methods can be further divided into BEV~\cite{yang2018pixor, yang2018hdnet} methods, discretization~\cite{zhou2018voxelnet, sindagi2019mvx, lang2019pointpillars} methods, and point-based~\cite{yang20203dssd} methods. Theoretically, VoteNet can be recognized as a region proposal-based method~\cite{guo2020deep}; meanwhile, it also shares the efficiency of point-based methods that can directly use point cloud data as input.

\vspace{3mm}
\noindent \textbf{Attention Mechanism.}
Initially, the attention mechanism intends to imitate the human vision system by focusing on the more relevant features to our targets rather than the whole scene containing some irrelevant context. Many methods have been introduced to estimate attention (weight) maps in order to re-weight the original feature map learned from the CNNs. As for image-related tasks, the attention map can be generated according to spatial~\cite{wang2018non,huang2019ccnet} or channel-related~\cite{hu2018squeeze, anwar2019real, fang2021attention} information, while some methods~\cite{fu2019dual,woo2018cbam} incorporate both of them for better information integration. In addition, point cloud networks tend to utilize a self-attention~\cite{vaswani2017attention} structure, which can estimate the long-range dependencies regardless of a specific order between the elements. In practice, we can leverage the basic form of self-attention to calculate either point-wise relations~\cite{xie2018attentional, feng2020point} or channel-wise affinities~\cite{qiu2021geometric} in a wide range of point cloud analysis problems~\cite{qiu2021pnp} such as classification, semantic segmentation, object detection, \etc. 

In this work, we adopt \emph{ten} standard attention modules covering the main types of existing designs used in both 2D images and 3D point clouds. By applying them to the backbone of VoteNet, we can comprehensively study the role of the attention mechanism in 3D point cloud object detection.

%% file: sections/approach.tex
\begin{figure}
\begin{center}
\includegraphics[width=0.93\columnwidth]{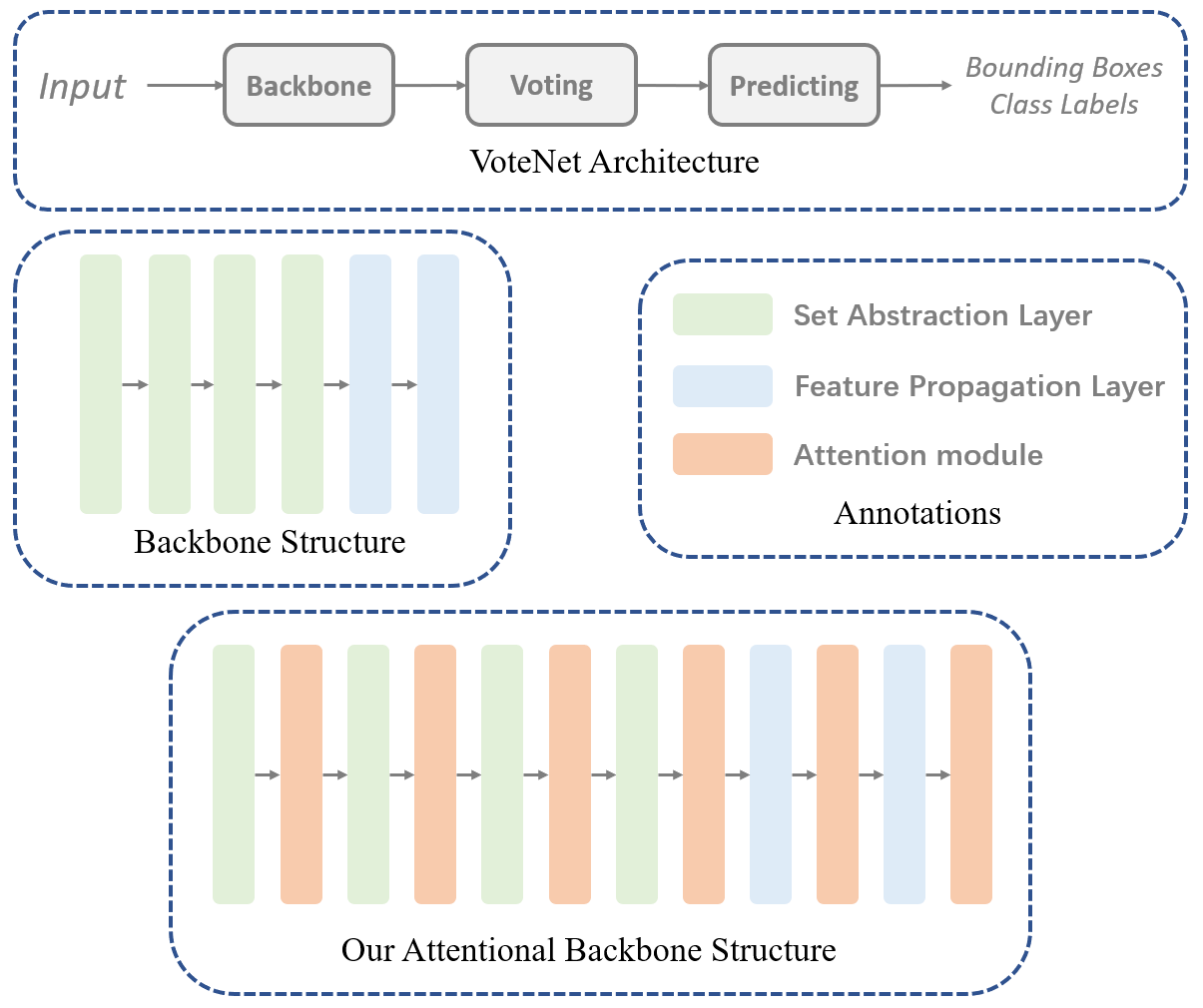}
\end{center}
\vspace{-5mm}
\caption{An overview of VoteNet~\cite{qi2019deep} pipeline including the structure of our attentional backbone.}
\label{fig:net}
\end{figure}

\section{Approach}
\label{sec:app}
As shown in Figure~\ref{fig:net}, the VoteNet pipeline consists of a backbone that learns features, a voting module estimating the object centers, as well as a predicting module regressing the bounding boxes and class labels. 

More concretely, the last two rows of Figure~\ref{fig:net} compare the detailed structures of VoteNet's original backbone and our attentional backbone. In general, the Set Abstraction (SA)~\cite{qi2017pointnet++} layer and the Feature Propagation (FP)~\cite{qi2017pointnet++} layer act as the encoder and decoder in the backbone, respectively. Following a regular usage in image-related CNNs, the attention module is placed after each encoder (SA) and decoder (FP) of the backbone. Moreover, to generate only a few seed points (\eg, 1024) from all input points (\eg, 20,000), we adopt the official implementation\footnote{\url{https://github.com/facebookresearch/votenet}} of VoteNet, which leverages four SA layers in a down-sampling manner while only two FP layers for up-sampling.

\begin{figure*}
\begin{center}
\includegraphics[width=\textwidth]{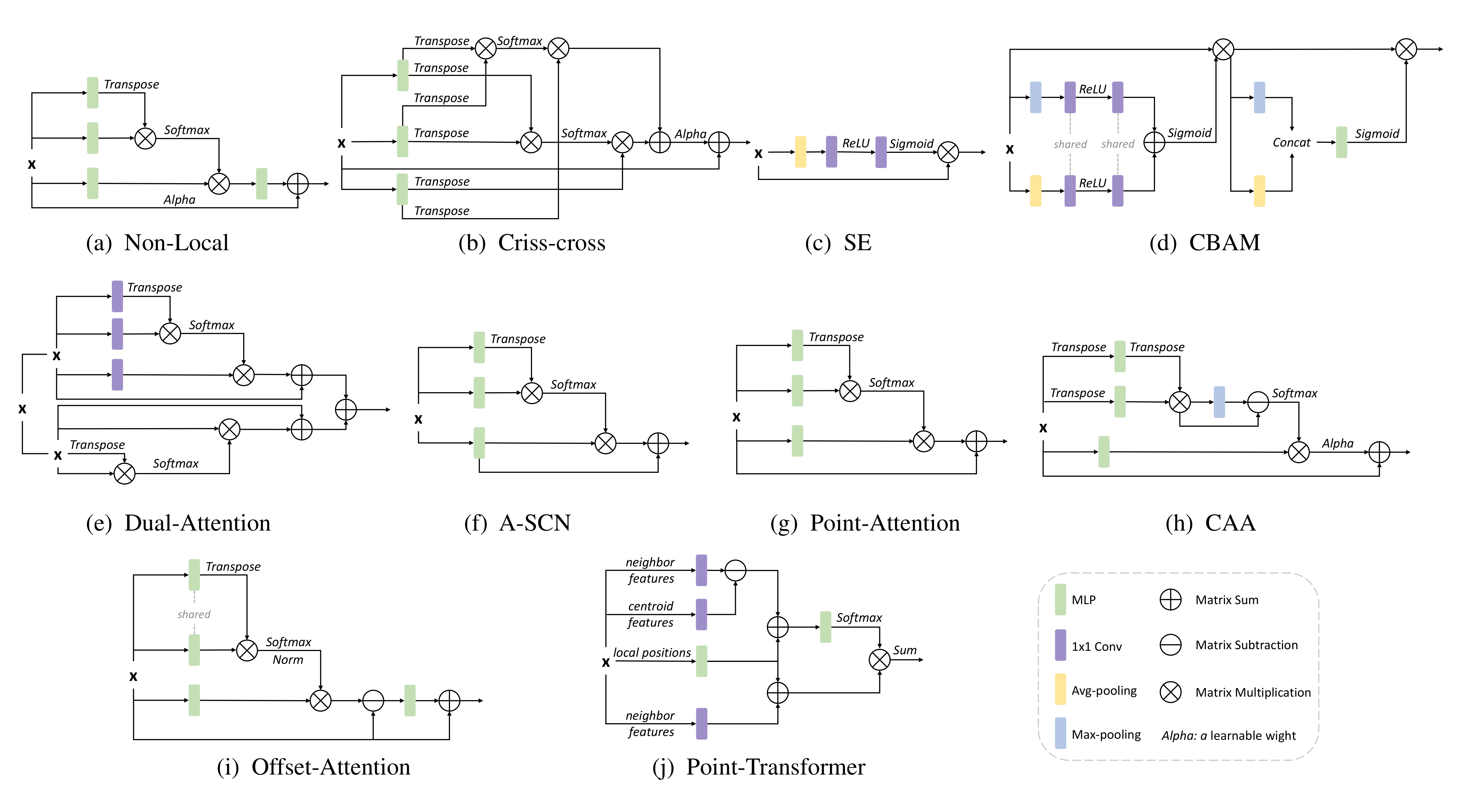}
\end{center}
\caption{The detailed structures of different attention modules that are investigated in this work. (2D attentions: Non-Local~\cite{wang2018non}, Criss-cross~\cite{huang2019ccnet}, SE~\cite{hu2018squeeze}, CBAM~\cite{woo2018cbam}, Dual-Attention~\cite{fu2019dual}; 3D attentions: A-SCN~\cite{xie2018attentional}, Point-Attention~\cite{feng2020point}, CAA~\cite{qiu2021geometric}, Offset-Attention~\cite{guo2021pct}, Point-Transformer~\cite{zhao2020point}.)}
\label{fig:attentions}
\end{figure*}
\subsection{2D Attentions}

Non-local~\cite{wang2018non} block is spatial attention that uses a weighted sum of features to represent a pixel. Particularly, the weights are estimated as the long-range dependencies (\ie, inner produces) between the pixels, by which the learned feature maps can be further enhanced with both local and global information.

Criss-cross attention~\cite{huang2019ccnet} is another spatial attention module, which exploits the criss-cross pixels to obtain the contextual information of a certain point effectively. Compared to the Non-local block, the criss-cross attention saves memory and achieves comparable performance in the meantime.

Squeeze and Excitation (SE) block~\cite{hu2018squeeze} is channel attention that intends to refine the features by exploiting the inter-dependencies between channels. At first, the SE block squeezes the spatial information and generates the channel-wise descriptors; then, it learns the weights for different channels by exciting the descriptors with convolutions and activations.

Convolutional Block Attention Module (CBAM)~\cite{woo2018cbam} is mixed attention that consists of two sequential attention blocks. To be specific, a channel attention block is leveraged to capture the channel-wise information based on the inter-channel relationship of features. In practice, it utilizes the sum of space-wise average-pooled and max-pooled descriptors to encode a channel attention map. In addition, the spatial attention block exploits the inter-spatial relationship of features for complementary context. Differently, the spatial attention block applies the average-pooling and max-pooling operations along the channels, then concatenates the pooled features to generate a spatial attention map.

Moreover, Dual-Attention~\cite{fu2019dual} is another mixed attention structure exploiting both channel-wise and spatial-wise long-range dependencies of features to enhance the discriminant ability. Notably, this structure utilizes a position attention module and a channel attention module in parallel. To be more specific, the position attention module can model the dependencies between positions in a feature map following a similar manner of the Non-local~\cite{wang2018non} block. In contrast, the channel attention module directly estimates the dependencies between the channels without any convolution. Finally, the two modules are aggregated using simple element-wise summation to generate the final feature map.

\begin{table*}
\begin{center}
\captionsetup{skip=3pt, font=normalsize}
\caption{The results of \emph{Average Precision} on \emph{SUN RGB-D}~\cite{song2015sun} dataset. (IoU threshold = 0.25)} 
\resizebox{0.9\textwidth}{!}{
\begin{tabular}{c|c| c c c c c c c c c c| c}
\Xhline{3\arrayrulewidth}
&\multirow{2}{*}{Method} &\multirow{2}{*}{bed} &\multirow{2}{*}{table} &\multirow{2}{*}{sofa} &\multirow{2}{*}{chair} &\multirow{2}{*}{toilet} &\multirow{2}{*}{desk} &\multirow{2}{*}{dresser} &night- &book- &bath- &\multirow{2}{*}{\textbf{mAP}}\\
&&&&&&&&&stand&shelf&tub&\\ \midrule
&\emph{baseline}~\cite{qi2019deep}& 83.3 & 49.8 & 64.1 & 74.1 & 89.3 & 23.8 & 26.4 & 60.7 & 30.9 & 72.8 & 57.5 \\ \midrule
\multirow{5}{*}{\rotatebox[origin=c]{90}{2D Attentions}}&Non-local~\cite{wang2018non}& 84.7 & \textbf{51.4} &62.9 &75.1 &89.4 &24.3 &28.8 &61.8 &28.0 &76.6 &58.3\\
&Criss-cross~\cite{huang2019ccnet}  &82.9 &49.8 &62.1 &74.1 &85.9 &24.2 &27.3 &60.2 &28.1 &67.2 &56.2 \\
&SE~\cite{hu2018squeeze} &84.2 &50.7 &\textbf{65.0} &\textbf{75.3} &\textbf{90.6} &\textbf{26.8} &32.3 &63.4 &31.6 &76.5 &\textbf{59.6} \\
&CBAM~\cite{woo2018cbam} &\textbf{84.8} &50.7 &64.1 &74.5 &90.4 &25.8 &\textbf{33.7} &\textbf{65.9} &28.8 &72.0 &59.1 \\
&Dual-attn~\cite{fu2019dual} &79.7 &44.5 &54.3 &67.4 &86.5 &18.6 &23.8 &45.8 &18.1 &67.1 &50.6 \\\midrule
\multirow{5}{*}{\rotatebox[origin=c]{90}{3D Attentions}}&A-SCN~\cite{xie2018attentional} &81.8 &48.9 &63.8 &74.0 &88.3 &24.5 &26.7 &57.5 &24.9 &65.4 &55.6 \\
&Point-attn~\cite{feng2020point} &84.4 &49.0 &61.9 &73.8 &87.4 &25.7 &24.6 &56.0 &28.2 &73.1 &56.4 \\
&CAA~\cite{qiu2021geometric} &83.7 &50.2 &63.4 &74.9 &89.7 &25.7 &30.6 &64.7 &27.5 &\textbf{77.6} &58.8 \\
&Point-trans~\cite{zhao2020point} &83.9 &50.4 &63.7 &75.2 &86.6 &26.3 &28.1 &62.5 &\textbf{35.8} &72.2 &58.5\\
&Offset-attn~\cite{guo2021pct} &82.8 &49.8 &60.5 &73.0 &86.5 &23.6 &27.1 &56.5 &25.6 &71.2 &55.7 \\
\Xhline{3\arrayrulewidth}
\end{tabular}
\label{tab:sun} 
}
\end{center} 
\end{table*}

\begin{table*}
\begin{center}
\captionsetup{skip=3pt, font=normalsize}
\caption{The results of \emph{Recall} on \emph{SUN RGB-D}~\cite{song2015sun} dataset. (IoU threshold = 0.25)} 
\resizebox{0.9\textwidth}{!}{
\begin{tabular}{c|c| c c c c c c c c c c| c}
\Xhline{3\arrayrulewidth}
&\multirow{2}{*}{Method} &\multirow{2}{*}{bed} &\multirow{2}{*}{table} &\multirow{2}{*}{sofa} &\multirow{2}{*}{chair} &\multirow{2}{*}{toilet} &\multirow{2}{*}{desk} &\multirow{2}{*}{dresser} &night- &book- &bath- &\multirow{2}{*}{\textbf{AR}}\\
&&&&&&&&&stand&shelf&tub&\\ \midrule
&\emph{baseline}~\cite{qi2019deep}&95.2 &85.5 &89.5 &86.7 &97.4 &78.8 &81.0 &87.8 &68.6 &90.4 &86.1 \\ \midrule
\multirow{5}{*}{\rotatebox[origin=c]{90}{2D Attentions}}&Non-local~\cite{wang2018non}&95.4 &84.6 &89.2 &87.2 &96.0 &79.7 &81.9 &\textbf{90.6} &63.2 &\textbf{92.3} &86.0\\
&Criss-cross~\cite{huang2019ccnet}  &93.4 &84.2 &89.0 &86.7 &94.7 &78.3 &82.4 &89.8 &66.2 &84.6 &84.9 \\
&SE~\cite{hu2018squeeze} &94.5 &\textbf{85.6} &89.2 &\textbf{86.9} &\textbf{99.3} &\textbf{80.4} &82.4 &89.8 &69.2 &90.4 &\textbf{86.8} \\
&CBAM~\cite{woo2018cbam} &\textbf{95.9} &84.7 &\textbf{90.1} &86.7 &97.4 &79.1 &\textbf{83.8} &90.2 &68.6 &86.5 &86.3 \\
&Dual-attn~\cite{fu2019dual} &92.1 &80.9 &86.1 &84.1 &95.4 &77.2 &79.2 &83.1 &66.2 &84.6 &82.9  \\\midrule
\multirow{5}{*}{\rotatebox[origin=c]{90}{3D Attentions}}&A-SCN~\cite{xie2018attentional} &94.1 &83.3 &88.4 &87.3 &96.7 &78.8 &77.3 &85.4 &67.6 &80.8 &84.0 \\
&Point-attn~\cite{feng2020point} &94.8 &83.6 &88.9 &86.3 &95.4 &78.7 &78.2 &88.2 &62.5 &86.5 &84.3 \\
&CAA~\cite{qiu2021geometric} &94.1 &84.7 &89.7 &86.8 &97.4 &79.3 &80.6 &89.8 &65.9 &90.4 &85.9 \\
&Point-trans~\cite{zhao2020point} &93.4 &84.5 &89.4 &86.1 &94.7 &77.4 &80.6 &89.4 &\textbf{71.9} &90.4 &85.8\\
&Offset-attn~\cite{guo2021pct} &94.1 &83.5 &87.8 &86.1 &97.4 &78.9 &78.2 &88.2 &64.9 &86.5 &84.6 \\
\Xhline{3\arrayrulewidth}
\end{tabular}
\label{tab:sun_recall} 
}
\end{center} 
\end{table*}

\subsection{3D Attentions}
Attentional ShapeContextNet (A-SCN)~\cite{xie2018attentional} introduces a self-attention-based module to exploit the shape context-driven features in the 3D point cloud. By comparing the query and key matrices, the attention map is estimated as the point-wise similarities. The output is then calculated as a matrix product between the attention map and the value matrix, together with an additional skip connection from the value matrix.

Point-Attention~\cite{feng2020point} module also follows the basic structure of self-attention to captures more shape-related features and long-range correlations from the point space of local point graphs. Additionally, it applies a skip-connection to strengthen the relationship between the input and output.

Channel Affinity Attention~\cite{qiu2021geometric} estimates the attention map between the channels by calculating the channel-wise affinities in a self-attention structure. Specifically, it utilizes a compact channel-wise comparator block and a channel affinity estimator block to compute the similarity matrix and affinity matrix.

Recently, inspiring by the success of transformers~\cite{khan2021transformers} in the 2D image, researchers also realize the transformer-based networks for point cloud analysis, in order to heavily exploit the attention mechanism as the basic point feature learning module. For example, Offset-Attention~\cite{guo2021pct} is proposed to estimate the offsets between the input and attention features, which are calculated from a self-attention structure. Mainly, Offset-Attention leverages the robustness of relative coordinates in transformations and the effectiveness of the Laplacian matrix in graph convolution. 

Moreover, Point Transformer~\cite{zhao2020point} is designed to take advantage of the local geometric relations between the center point and its neighbors. Using basic MLP operations, the Point Transformer block can effectively aggregate a local feature for each point based on the learned attention weights for its neighbors. With the help of rich local and geometric context, this method achieves outstanding performances in both point cloud classification and segmentation tasks.

%% file: sections/exp.tex
\begin{table*}
\begin{center}
\captionsetup{skip=3pt, font=normalsize}
\caption{The results of \emph{Average Precision} on \emph{ScanNetV2}~\cite{dai2017scannet} dataset. (IoU threshold = 0.25)} 
\resizebox{\textwidth}{!}{
\begin{tabular}{c|c| c c c c c c c c c c c c c c c c c c| c}
\Xhline{3\arrayrulewidth}
&\multirow{2}{*}{Method} &\multirow{2}{*}{cabinet} &\multirow{2}{*}{bed} &\multirow{2}{*}{chair} &\multirow{2}{*}{sofa} &\multirow{2}{*}{table} &\multirow{2}{*}{door} &\multirow{2}{*}{window} &book- &\multirow{2}{*}{picture} &\multirow{2}{*}{counter} &\multirow{2}{*}{desk} &\multirow{2}{*}{curtain} &refri- &shower- &\multirow{2}{*}{toilet} &\multirow{2}{*}{sink} &\multirow{2}{*}{bathtub} &garba- &\multirow{2}{*}{\textbf{mAP}} \\
&&&&&&&&&shelf&&&&&gerator&curtain&&&&gebin&\\\midrule
&\emph{baseline}~\cite{qi2019deep} &38.5 &\textbf{88.7} &87.8 &\textbf{90.4} &58.9 &45.0 &36.7 &44.8 &4.9 &50.0 &61.4 &39.1 &50.4 &59.1 &97.3 &48.7 &91.3 &39.0 &57.3 \\ \midrule
\multirow{5}{*}{\rotatebox[origin=c]{90}{2D Attentions}} &Non-local~\cite{wang2018non} &33.3 &86.6 &87.5 &85.4 &58.7 &42.4 &33.4 &47.9 &3.6 &50.9 &66.5 &40.3 &51.4 &\textbf{65.9} &96.9 &\textbf{57.2} &\textbf{92.9} &34.4 &57.5\\
&Criss-cross~\cite{huang2019ccnet}  &37.7 &86.7 &86.3 &85.2 &60.3 &40.8 &34.6 &45.1 &5.0 &57.8 &\textbf{71.5} &40.2 &43.9 &61.1 &94.0 &47.4 &88.9 &34.8 &56.7 \\
&SE~\cite{hu2018squeeze} &35.3 &88.6 &87.5 &86.7 &59.4 &44.7 &35.3 &\textbf{57.5} &5.6 &49.6 &70.8 &\textbf{47.1} &49.2 &61.9 &95.7 &50.4 &\textbf{92.9} &36.3 &58.6 \\
&CBAM~\cite{woo2018cbam} &\textbf{39.2} &\textbf{88.7} &87.9 &89.5 &60.1 &\textbf{48.2} &38.5 &49.4 &5.3 &51.9 &69.6 &42.5 &\textbf{54.3} &61.7 &93.3 &49.0 &88.4 &38.6 &58.7 \\
&Dual-attn~\cite{fu2019dual} &34.7 &88.2 &86.5 &84.4 &56.4 &42.2 &27.0 &41.3 &2.6 &51.7 &66.1 &37.2 &46.3 &56.6 &98.3 &46.2 &85.4 &33.4 &54.7 \\\midrule
\multirow{5}{*}{\rotatebox[origin=c]{90}{3D Attentions}} &A-SCN~\cite{xie2018attentional} &37.3 &85.7 &88.2 &87.9 &58.2 &41.3 &31.8 &46.8 &3.5 &50.9 &67.9 &35.8 &49.6 &61.3 &96.6 &53.2 &83.9 &37.8 &56.5 \\
&Point-attn~\cite{feng2020point} &31.8 &87.4 &84.0 &88.4 &58.5 &38.2 &31.5 &41.2 &2.2 &61.2 &69.1 &29.6 &50.7 &49.5 &97.3 &46.6 &83.9 &33.4 &54.7 \\
&CAA~\cite{qiu2021geometric} &36.4 &88.5 &\textbf{88.7} &89.7 &60.0 &44.5 &38.6 &48.4 &4.4 &49.3 &69.8 &39.0 &43.1 &60.4 &94.3 &53.0 &91.3 &37.2 &57.6 \\
&Point-trans~\cite{zhao2020point} &39.0 &84.5 &88.3 &88.3 &\textbf{63.0} &44.5 &\textbf{39.5} &53.4 &6.6 &52.6 &70.2 &41.6 &46.8 &63.1 &\textbf{97.4} &48.4 &91.6 &\textbf{44.9} &\textbf{59.1}\\
&Offset-attn~\cite{guo2021pct} &38.0 &88.1 &87.2 &89.9 &58.5 &43.2 &27.5 &50.2 &\textbf{6.8} &\textbf{59.6} &69.9 &39.5 &50.6 &61.5 &95.8 &51.1 &87.2 &38.9 &58.0 \\
\Xhline{3\arrayrulewidth}
\end{tabular}
\label{tab:scan} 
}
\end{center} 
\end{table*}

\begin{table*}
\begin{center}
\captionsetup{skip=3pt, font=normalsize}
\caption{The results of \emph{Recall} on \emph{ScanNetV2}~\cite{dai2017scannet} dataset. (IoU threshold = 0.25)} 
\resizebox{\textwidth}{!}{
\begin{tabular}{c|c| c c c c c c c c c c c c c c c c c c| c}
\Xhline{3\arrayrulewidth}
&\multirow{2}{*}{Method} &\multirow{2}{*}{cabinet} &\multirow{2}{*}{bed} &\multirow{2}{*}{chair} &\multirow{2}{*}{sofa} &\multirow{2}{*}{table} &\multirow{2}{*}{door} &\multirow{2}{*}{window} &book- &\multirow{2}{*}{picture} &\multirow{2}{*}{counter} &\multirow{2}{*}{desk} &\multirow{2}{*}{curtain} &refri- &shower- &\multirow{2}{*}{toilet} &\multirow{2}{*}{sink} &\multirow{2}{*}{bathtub} &garba- &\multirow{2}{*}{\textbf{AR}} \\
&&&&&&&&&shelf&&&&&gerator&curtain&&&&gebin&\\\midrule
&\emph{baseline}~\cite{qi2019deep} &76.3 &\textbf{95.1} &91.9 &\textbf{99.0} &82.0 &\textbf{72.4} &63.8 &84.4 &23.0 &84.6 &93.7 &71.6 &93.0 &78.6 &98.3 &64.3 &\textbf{96.8} &\textbf{70.6} &80.0 \\ \midrule
\multirow{5}{*}{\rotatebox[origin=c]{90}{2D Attentions}} &Non-local~\cite{wang2018non} &74.2 &93.8 &91.8 &97.9 &84.0 &71.3 &60.6 &81.8 &19.4 &82.7 &94.5 &\textbf{79.1} &93.0 &\textbf{96.4} &98.3 &\textbf{74.5} &\textbf{96.8} &65.7 &80.9\\
&Criss-cross~\cite{huang2019ccnet}  &73.9 &\textbf{95.1} &91.3 &97.9 &82.9 &68.5 &61.0 &85.7 &18.0 &\textbf{86.5} &93.7 &73.1 &94.7 &78.6 &94.8 &67.3 &93.5 &67.4 &79.1 \\
&SE~\cite{hu2018squeeze} &72.6 &\textbf{95.1} &92.0 &96.9 &82.9 &71.3 &64.5 &85.7 &21.2 &80.8 &\textbf{96.1} &77.6 &91.2 &92.9 &96.6 &63.3 &\textbf{96.8} &67.7 &80.3 \\
&CBAM~\cite{woo2018cbam} &\textbf{77.2} &\textbf{95.1} &92.1 &97.9 &\textbf{84.3} &70.7 &\textbf{69.1} &\textbf{87.0} &19.8 &82.7 &94.5 &76.1 &96.5 &\textbf{96.4} &96.6 &65.3 &90.3 &65.8 &\textbf{81.0} \\
&Dual-attn~\cite{fu2019dual} &73.4 &\textbf{95.1} &92.3 &97.9 &81.7 &71.3 &56.0 &85.7 &18.0 &\textbf{86.5} &93.7 &74.6 &\textbf{98.2} &92.9 &\textbf{100.0} &65.3 &93.5 &67.0 &80.2 \\\midrule
\multirow{5}{*}{\rotatebox[origin=c]{90}{3D Attentions}} &A-SCN~\cite{xie2018attentional} &75.8 &\textbf{95.1} &93.0 &97.9 &83.7 &71.5 &62.8 &84.4 &19.4 &82.7 &93.7 &71.6 &\textbf{98.2} &92.9 &\textbf{100.0} &71.4 &90.3 &66.8 &80.6 \\
&Point-attn~\cite{feng2020point} &71.2 &93.8 &90.4 &97.9 &82.6 &70.0 &61.7 &84.4 &17.6 &84.6 &95.3 &74.6 &96.5 &85.7 &\textbf{100.0} &66.3 &90.3 &66.6 &79.4 \\
&CAA~\cite{qiu2021geometric} &74.2 &93.8 &92.1 &\textbf{99.0} &84.0 &71.5 &65.6 &83.1 &21.2 &84.6 &95.3 &73.1 &94.7 &92.9 &94.8 &67.3 &\textbf{96.8} &66.2 &80.6 \\
&Point-trans~\cite{zhao2020point} &75.8 &\textbf{95.1} &\textbf{92.5} &95.9 &82.6 &70.0 &64.9 &83.1 &\textbf{25.2} &84.6 &94.5 &77.6 &91.2 &89.3 &98.3 &65.3 &93.5 &66.8 &80.3\\
&Offset-attn~\cite{guo2021pct} &73.9 &\textbf{95.1} &92.1 &97.9 &81.4 &71.7 &56.7 &80.5 &20.7 &\textbf{86.5} &\textbf{96.1} &73.1 &\textbf{98.2} &92.9 &98.3 &67.3 &90.3 &65.5 &79.9 \\
\Xhline{3\arrayrulewidth}
\end{tabular}
\label{tab:scan_recall} 
}
\end{center} 
\end{table*}

\section{Experiments}

\subsection{Datasets}
We evaluate the performances of different attention modules on two datasets, \emph{SUN RGB-D}~\cite{song2015sun} and \emph{ScanNetV2}~\cite{dai2017scannet}, which are both captured from the indoor scenes in real-world using RBG-D cameras. To be more specific:
\begin{itemize}
    \item \textbf{SUN RGB-D:} There are 5,285 training and 5,050 testing RGB-D images in the dataset, where each object is precisely annotated with a bounding box and one of 35 semantic classes. According to the provided camera parameters, the original data and annotated bounding boxes can be projected as 3D point clouds. Following a widely used experimental setting~\cite{qi2019deep}, we only use the 3D coordinates as input and report the average precision (AP) and recall of the ten most common classes, together with the overall metrics of mean average precision (mAP) and average recall (AR).
    \item \textbf{ScanNetV2:} The original dataset contains the reconstructed meshes from 18 object categories, where 1,201 samples are for training and 312 samples are in the validation set. The point cloud data is sampled from the vertices of reconstructed meshes. Besides the usage in segmentation, we input the 3D coordinates and predict the bounding box and category of each object in the same evaluating protocol as~\cite{qi2019deep}.
\end{itemize}

\subsection{Implementation}

In general, all attention modules in our work are adopted and slightly modified from the available official implementations. For 2D attentions, we regard the spatial space of 2D image as the point space of point cloud, following relation of $H \times W = N \times 1;$ where $H$ and $W$ are the height and width of an image while $N$ is the number of points. Moreover, to achieve a stable training process in 3D cases, we replace the original $1 \times 1$ convolutions in 2D attentions with the MLP operations if necessary. For fair comparisons, the reduction factor in all attention modules is empirically set as 8. As for the recent Point Transformer~\cite{zhao2020point} whose code is not released yet, we reproduce the structure according to the related descriptions in the paper.

The implementations are realized by PyTorch and Python platforms on a single Tesla-P100 GPU using CUDA and Linux operating system. All the experiments adopt the similar training settings such as the learning rate of 0.001, the batch size of 8, and a total of 180 training epochs. Following the default configurations in~\cite{qi2019deep}, in \emph{SUN RGB-D}~\cite{song2015sun} dataset, the number of input points is 20,000; while in \emph{ScanNetV2}~\cite{dai2017scannet} dataset, the input size is 40,000.  Coupled with this paper, we will release the source code of all deployed attention modules.

\subsection{Experimental Results}
\label{sec:results}
By applying different attention modules in our attentional backbone, we conduct the experiments of 3D point cloud object detection on \emph{SUN RGB-D}~\cite{song2015sun} dataset and \emph{ScanNetV2}~\cite{dai2017scannet} dataset, respectively.

Table~\ref{tab:sun} and~\ref{tab:sun_recall} present the detailed detection results in \emph{SUN RGB-D}~\cite{song2015sun} dataset under the metrics of average precision (AP), mean average precision (mAP), recall and average recall (AR). Although we can notice the different effects by using different attention modules, as shown in Table~\ref{tab:sun}, the SE~\cite{liu2019point2sequence} method realizes the best overall result (59.6\% mAP) among all tested attention modules and significantly exceeds the baseline's result (57.5\% mAP) by 2.1\%. In terms of each category's AP, the SE~\cite{liu2019point2sequence} method achieves the highest values in four out of ten object categories. Meanwhile, the results in Table~\ref{tab:sun_recall} can also verify the outstanding performance of SE~\cite{hu2018squeeze} under the metrics of recall. In comparison to the complicated self-attention methods, a compact attention structure like in SE~\cite{liu2019point2sequence} or CBAM~\cite{woo2018cbam} is able to benefit the point cloud detection task effectively and efficiently. Moreover, we observe that the channel-related information plays a crucial role in the attention mechanism of the point cloud, since the effectiveness of SE~\cite{hu2018squeeze}, CBAM~\cite{woo2018cbam} and CAA~\cite{qiu2021geometric} are more prominent than the spatial attention modules.

\begin{table}
\begin{center}
\captionsetup{skip=3pt, font=normalsize}
\caption{Overall evaluations of 3D point cloud object detection results using \emph{SUN RGB-D}~\cite{song2015sun} and \emph{ScanNetV2}~\cite{dai2017scannet} datasets. (\enquote{\textbf{mAP}}: mean average precision; \enquote{\textbf{AR}}: average recall; the value behind \enquote{@} denotes the IoU threshold.)} 
\resizebox{0.95\columnwidth}{!}{
\begin{tabular}{c|c|cc|cc}
\Xhline{3\arrayrulewidth}
& \multirow{2}{*}{Method} & \multicolumn{2}{c|}{\emph{SUN RGB-D}~\cite{song2015sun}} &\multicolumn{2}{c}{\emph{ScanNetV2}~\cite{dai2017scannet}} \\
&&mAP@0.5 &AR@0.5 &mAP@0.5 &AR@0.5\\\midrule
& \emph{baseline}~\cite{qi2019deep} &33.1 &51.1 &33.7 &49.9 \\ \midrule
\multirow{5}{*}{\rotatebox[origin=c]{90}{2D Attentions}} &Non-local~\cite{wang2018non} &31.4 &49.7 &34.6 &49.5\\
& Criss-cross~\cite{huang2019ccnet}  &33.1 &50.0 &33.8 &49.2 \\
& SE~\cite{hu2018squeeze} &34.5 &52.1 &35.8 &51.4 \\
& CBAM~\cite{woo2018cbam} &\textbf{34.9} &\textbf{53.1} &37.1 &52.5 \\
& Dual-attn~\cite{fu2019dual} &24.4 &42.1 &30.2 &47.2 \\\midrule
\multirow{5}{*}{\rotatebox[origin=c]{90}{3D Attentions}} &A-SCN~\cite{xie2018attentional} &30.1 &48.2 &33.1 &48.7 \\
& Point-attn~\cite{feng2020point} &32.2 &49.7 &30.8 &46.7 \\
& CAA~\cite{qiu2021geometric} &33.3 &51.4 &35.1 &50.4 \\
& Point-trans~\cite{zhao2020point} &34.3 &51.3 &\textbf{38.0} &\textbf{53.5}\\
& Offset-attn~\cite{guo2021pct} &30.6 &48.2 &36.0 &50.4 \\
\Xhline{3\arrayrulewidth}
\end{tabular}
\label{tab:overall} 
}
\vspace{-3mm}
\end{center} 
\end{table}


In Table \ref{tab:scan} and~\ref{tab:scan_recall}, we further compare the performances of different attention modules under a challenging condition of more input points and detected object categories using \emph{ScanNetV2}~\cite{dai2017scannet} dataset. Even though SE~\cite{hu2018squeeze} and CBAM~\cite{woo2018cbam} can still provide relatively good performances evaluated under either precision- or recall-related metrics, the Point Transformer~\cite{zhao2020point} achieves the best overall result (59.1\% mAP) among all the ten tested attention modules. Mainly, the advantages of Point Transformer~\cite{zhao2020point} can be concluded from two sides: on the one hand, it incorporates more local context for each point rather than a single feature representation learned from a shared MLP; on the other hand, the attention map is estimated from the geometric relations in 3D space, while most of the rest methods only calculate the dependencies from the feature space. 

\subsection{Overall Evaluations}
In addition to the average precision and recall evaluated under an IoU threshold of 0.25 in Section~\ref{sec:results}, we provide the overall evaluations, mean average precision (mAP) and average recall (AR), under an IoU threshold of 0.5.

In general, Table~\ref{tab:overall} shows the similar results as in Table~\ref{tab:sun} and~\ref{tab:sun_recall}, where the SE~\cite{hu2018squeeze} and CBAM~\cite{woo2018cbam} methods achieve better mAP and AR scores under both the IoU thresholds of 0.25 and 0.5 in \emph{SUN RGB-D} dataset. Moreover, it is worth noting that when the IoU threshold is set at 0.5, Point Transformer~\cite{zhao2020point} achieves the improvements of 4.3\% mAP and 3.6\% AR compared to the baseline results, since it can integrate more local information for the object detection task in dense point cloud scenes in \emph{ScanNetV2}~\cite{dai2017scannet} dataset. 

Apart from VoteNet~\cite{qi2019deep}, we conduct more experiments by utilizing our attentional backbone in BoxNet~\cite{qi2019deep} and MLCVNet~\cite{xie2020mlcvnet}. In general, the attentions show similar effects with
them as when tested with VoteNet. More experimental data can be found in the supplementary material.

\begin{table}
\begin{center}
\captionsetup{skip=3pt, font=normalsize}
\caption{Model complexity of VoteNet~\cite{qi2019deep} with different attention modules, evaluated on \emph{ScanNetV2}~\cite{dai2017scannet} dataset. (\enquote{$^*$}: counted by the first attention module in the backbone.)} 
\resizebox{0.95\columnwidth}{!}{
\begin{tabular}{c|c|cccc}
\Xhline{3\arrayrulewidth}
& \multirow{2}{*}{Method} &\textbf{model size} &\textbf{training time} &\textbf{inference time} &\textbf{\# parameters} \\
& &(MB)&(s/epoch)&(s/epoch)&($\times10^3$/attention$^*$)\\\midrule
& \emph{baseline}~\cite{qi2019deep} &11.0 &43.8 &35.0 &-  \\ \midrule
\multirow{5}{*}{\rotatebox[origin=c]{90}{2D Attentions}} &Non-local~\cite{wang2018non} &13.0 &48.2 &35.9 &8.5\\
& Criss-cross~\cite{huang2019ccnet} &16.0 &54.6 &35.2 &20.8 \\
& SE~\cite{hu2018squeeze} &11.9 &44.2 &35.1 &4.1 \\
& CBAM~\cite{woo2018cbam} &11.5 &45.7 &36.4 &4.1 \\
& Dual-attn~\cite{fu2019dual} &15.9 &50.6 &36.7 &21.0 \\\midrule
\multirow{5}{*}{\rotatebox[origin=c]{90}{3D Attentions}} &A-SCN~\cite{xie2018attentional} &16.0 &48.5 &35.9 &20.8 \\
& Point-attn~\cite{feng2020point} &16.0 &48.6 &35.6 &20.8 \\
& CAA~\cite{qiu2021geometric} &34.7 &47.2 &36.7 &106.6 \\
& Point-trans~\cite{zhao2020point} &25.8 &88.1 &38.7 &100.1\\
& Offset-attn~\cite{guo2021pct} &19.5 &50.1 &35.3 &35.6 \\
\Xhline{3\arrayrulewidth}
\end{tabular}
\label{tab:complexity} 
}
\vspace{-3mm}
\end{center} 
\end{table}

\begin{figure*}[ht]
\centering
\begin{subfigure}{.24\textwidth}
  \centering
  \includegraphics[width=\linewidth]{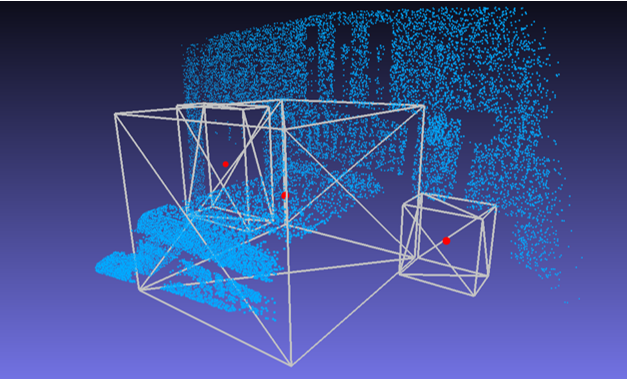}
  \caption{Input~\cite{song2015sun}}
\end{subfigure}
\vspace{1mm}
\begin{subfigure}{.24\textwidth}
  \centering
  \includegraphics[width=\linewidth]{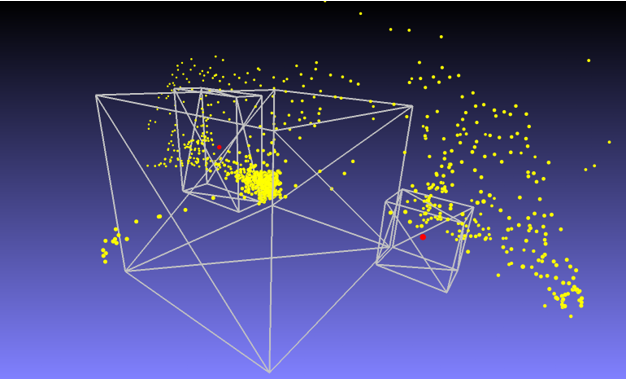}  
  \caption{Baseline~\cite{qi2019deep}}
\end{subfigure}
\begin{subfigure}{.24\textwidth}
  \centering
  \includegraphics[width=\linewidth]{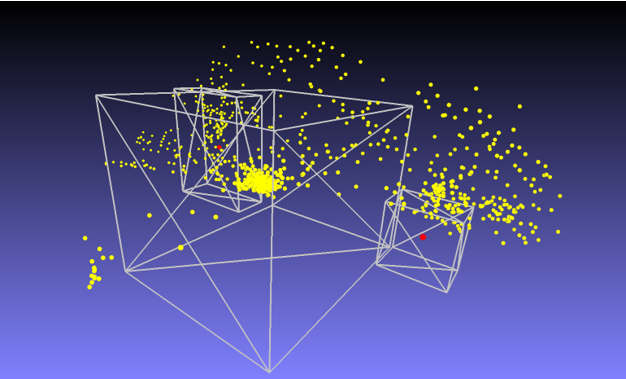}  
  \caption{Non-local~\cite{wang2018non}}
\end{subfigure}
\begin{subfigure}{.24\textwidth}
  \centering
  \includegraphics[width=\linewidth]{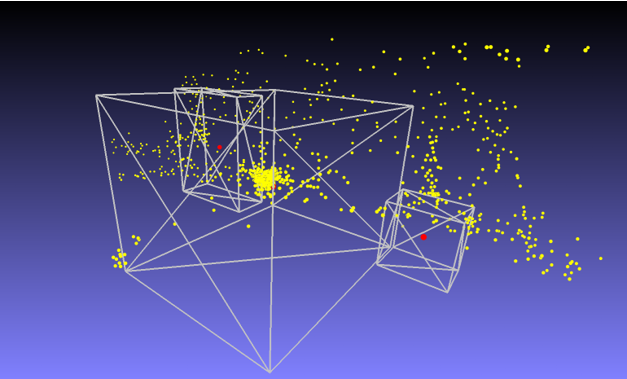}  
  \caption{Criss-cross~\cite{huang2019ccnet}}
\end{subfigure}
\hfil
\begin{subfigure}{.24\textwidth}
  \centering
  \includegraphics[width=\linewidth]{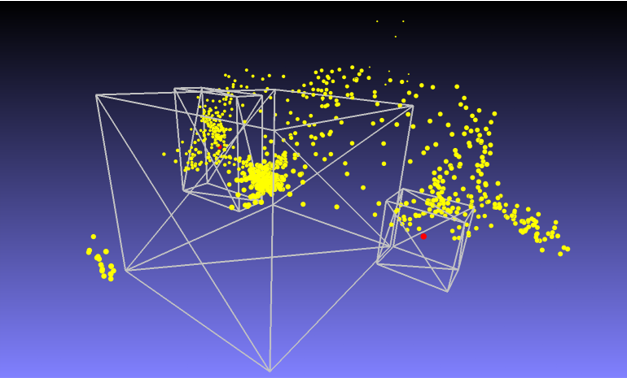}  
  \caption{SE~\cite{hu2018squeeze}}
\end{subfigure}
\vspace{1mm}
\begin{subfigure}{.24\textwidth}
  \centering
  \includegraphics[width=\linewidth]{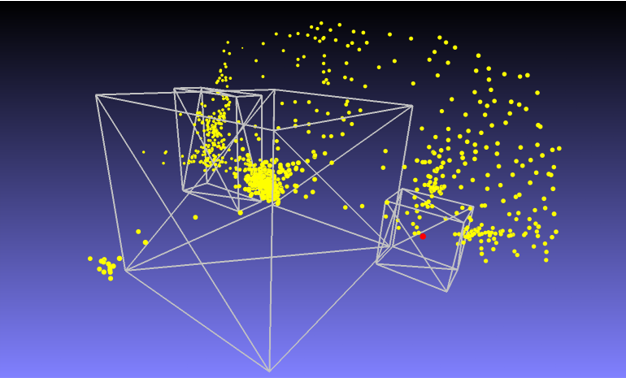}  
  \caption{CBAM~\cite{woo2018cbam}}
\end{subfigure}
\begin{subfigure}{.24\textwidth}
  \centering
  \includegraphics[width=\linewidth]{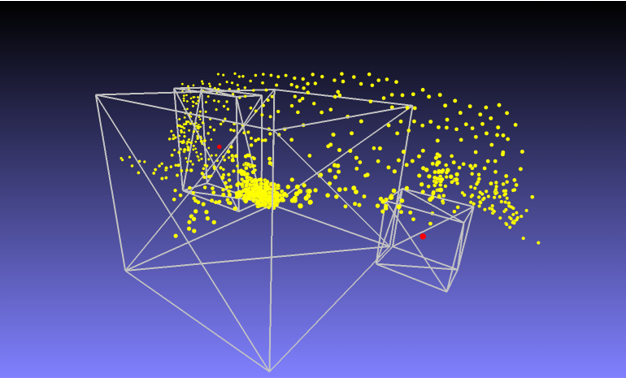}  
  \caption{Dual-Attention~\cite{fu2019dual}}
\end{subfigure}
\begin{subfigure}{.24\textwidth}
  \centering
  \includegraphics[width=\linewidth]{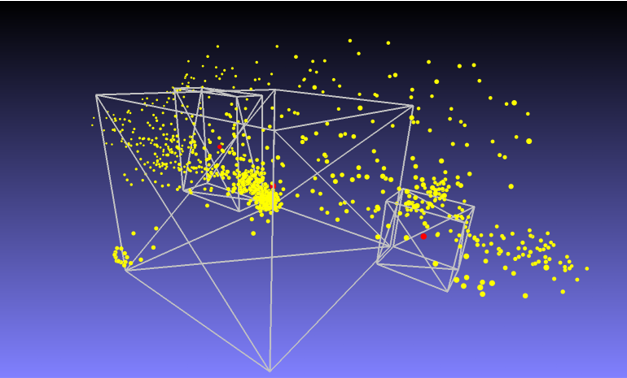}  
  \caption{A-SCN~\cite{xie2018attentional}}
\end{subfigure}
\hfil
\begin{subfigure}{.24\textwidth}
  \centering
  \includegraphics[width=\linewidth]{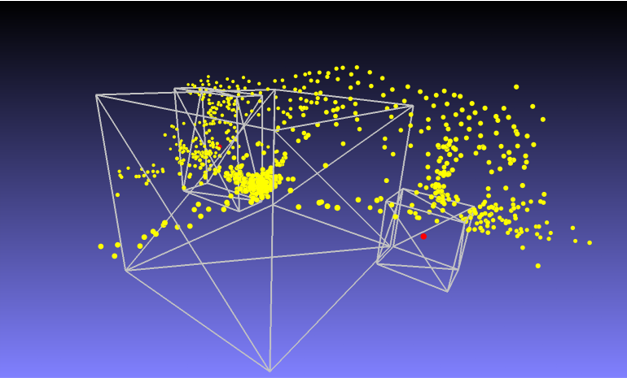}  
  \caption{Point-Attention~\cite{feng2020point}}
\end{subfigure}
\begin{subfigure}{.24\textwidth}
  \centering
  \includegraphics[width=\linewidth]{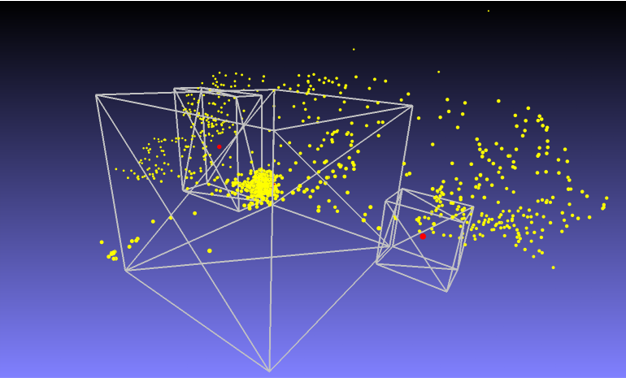}  
  \caption{CAA~\cite{qiu2021geometric}}
\end{subfigure}
\begin{subfigure}{.24\textwidth}
  \centering
  \includegraphics[width=\linewidth]{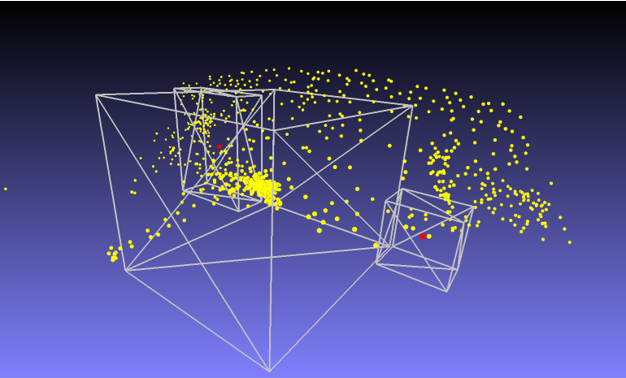}  
  \caption{Offset-Attention~\cite{guo2021pct}}
\end{subfigure}
\begin{subfigure}{.24\textwidth}
  \centering
  \includegraphics[width=\linewidth]{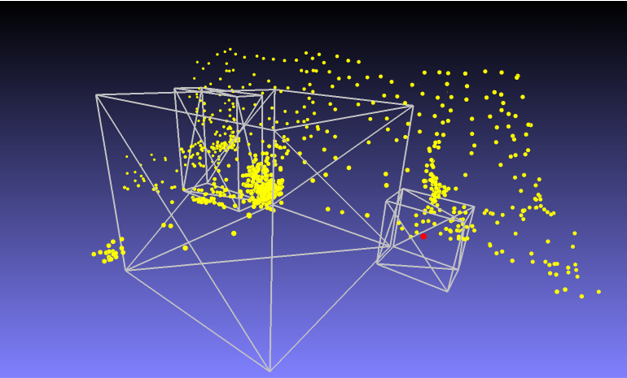}  
  \caption{Point-Transformer~\cite{zhao2020point}}
\end{subfigure}
\caption{Visualizations of the votes generated from VoteNet~\cite{qi2019deep} backbone by leveraging different attention modules. The input point cloud scene (from \emph{SUN RGB-D}~\cite{song2015sun} dataset) contains 3 objects for detection, where the bounding boxes (ground-truths) are drawn in white frames. Theoretically, the generated votes (yellow points) are expected to be around the centroids (red points) of detected objects as many as possible.}
\vspace{-3mm}
\label{fig:vis}
\end{figure*}

\subsection{Model Complexity}
Table~\ref{tab:complexity} presents some reference data regarding a model's complexity. By adding different attention modules in the backbone of VoteNet~\cite{qi2019deep}, the model size of whole network may increase more or less depending on the number of parameters used in each attention module. Although some attentions, \eg, Point Transformer~\cite{zhao2020point} and CAA~\cite{qiu2021geometric}, can achieve relatively higher performances, they require more computational resources such as longer training time or larger memory consumption. In terms of the inference time, all tested models can perform at a similar level. However, as Point Transformer~\cite{zhao2020point} needs an additional local neighbor searching operation compared to other methods, the efficiency of its inference process will be affected. Alternatively, we may integrate more global perception in both spatial and channel domains as in CBAM~\cite{woo2018cbam} and SE~\cite{hu2018squeeze}, to better balance the effectiveness and efficiency.

\subsection{Visualization}
Recall that in the VoteNet pipeline, the most critical usage of its backbone is to generate the votes (yellow points) that are expected to approach the centroids (red points) of detected objects. Therefore, the generated votes can intuitively indicate the quality of the backbone's output.

To this end, in Figure~\ref{fig:vis}, we compare the votes that are generated from different attentional backbones. In the sub-figure of the baseline, we can see that the votes can be easily attached to the most significant object's (the middle one) centroid, while there are fewer votes around the centroids of two smaller objects. As for the sub-figure of the SE method, it can be clearly observed that more votes have been centralized at the centroids of two smaller objects (especially for the left object), providing more confident estimations on the detected objects bounding boxes. To further visualize the effects of different attention modules, in the supplementary material, we also compare the point features learned from our attentional backbone.

\section{Insights}
From the experimental results and our analysis, we obtain several interesting
observations and insights of attention mechanism in 3D point cloud object detection:
\begin{itemize}
  \item [1)]
  The self-attention modules are not preferable in processing 3D point cloud data. On the one hand,  the fashion of self-attention needs high computational resources. On the other hand, the effectiveness of point-wise long-range dependencies used in self-attention modules is relatively limited as such an operation may cause some redundancies in representing the large-scale 3D point cloud data. 
  \item [2)]
  The compact attention structures like SE~\cite{hu2018squeeze} and CBAM~\cite{woo2018cbam} enable the effectiveness and efficiency of 3D point cloud feature refinement. This is achieved by capturing the global perception from a broad perspective in feature space.   
  \item [3)]
  Comparing the spatial attention module with the channel-attention module, we found that the channel-related information is more important than spatial information when embedded into the attention modules for point cloud feature representations.
  \item [4)]
  As reflected from the Point Transformer~\cite{zhao2020point}'s results, incorporating more local context could better represent the complex point cloud scenes, thus leading to better 3D point cloud object detection performance.
\end{itemize}

%% file: sections/conclusion.tex
\section{Conclusion}
This paper proposes an attentional backbone used in the VoteNet pipeline for 3D point cloud object detection. By integrating different standard 2D and 3D attentions modules, we compare their effects via various metrics and datasets. Based on the experiments and visualization, we summarize the effects of the attention mechanism in 3D point cloud object detection. Moreover, we provide our insights on how to effectively leverage the attention mechanism for point cloud feature representation. In addition to presenting a benchmark evaluating the performances of different attention modules, we expect our preliminary findings to help future research in designing reliable and transparent attention structures for more point cloud analyzing works.